\documentclass[letterpaper, 10 pt, conference]{ieeeconf}  

\IEEEoverridecommandlockouts                              

\usepackage{hyperref}       
\usepackage{url}            
\usepackage{booktabs}       
\usepackage{amsfonts}       
\usepackage{nicefrac}       
\usepackage{microtype}      
\usepackage{xcolor}         
\let\labelindent\relax
\usepackage{enumitem}
\usepackage{amsmath}
\usepackage{multirow}
\usepackage{booktabs}

\usepackage{array}
\usepackage{graphicx}
\usepackage{colortbl}
\usepackage{marvosym}

 \usepackage[ruled,vlined]{algorithm2e}
\usepackage{algorithmic}

\usepackage{adjustbox}
\usepackage{caption}
\title{\LARGE \bf
TransDiffuser: Diverse Trajectory Generation with Decorrelated Multi-modal Representation for End-to-end Autonomous Driving
}

\author{
Xuefeng Jiang$^{1,2}$, Yuan Ma$^{1,3}$, Pengxiang Li$^{1,3}$, Leimeng Xu$^{1}$, Xin Wen$^{1}$, \\ Kun Zhan$^{1\dagger}$, Zhongpu Xia$^{1}$, Peng Jia$^{1}$, Xianpeng Lang$^{1}$, Sheng Sun$^{2\dagger}$
\thanks{$^{1}$~LiAuto Inc}%
\thanks{$^{2}$~Institute of Computing Technology, Chinese Academy of Sciences}%
\thanks{$^{3}$~School of Vehicle and Mobility, Tsinghua University}%
\thanks{$^{\dagger}$~Corresponding authors.}%
\thanks{\Letter: jiangxuefeng21b@ict.ac.cn, zhankun@lixiang.com}
}

\begin{document}

\maketitle
\thispagestyle{empty}
\pagestyle{empty}

\begin{abstract}

In recent years, diffusion models have demonstrated remarkable potential across diverse domains, from vision generation to language modeling.
Transferring its generative capabilities to modern end-to-end autonomous driving systems has also emerged as a promising direction.
However, existing diffusion-based trajectory generative models often exhibit mode collapse where different random noises converge to similar trajectories after the denoising process.
Therefore, state-of-the-art models often rely on anchored trajectories from pre-defined trajectory vocabulary or scene priors in the training set to mitigate collapse and enrich the diversity of generated trajectories, but such inductive bias are not available in real-world deployment, which can be challenged when generalizing to unseen scenarios.
In this work, we investigate the possibility of effectively tackling the mode collapse challenge without the assumption of predefined trajectory vocabulary or pre-computed scene priors. 
Specifically, we propose TransDiffuser, an encoder-decoder based generative trajectory planning model, where the encoded scene information and motion states serve as the multi-modal conditional input of the denoising decoder.
Different from existing approaches, we exploit a simple yet effective multi-modal representation decorrelation optimization mechanism during the denoising process to enrich the latent representation space which better guides the downstream generation.
Without any pre-defined trajectory anchors or pre-computed scene priors, TransDiffuser achieves the PDMS of 94.85 on the closed-loop planning-oriented benchmark NAVSIM, surpassing previous state-of-the-art methods. 
Qualitative evaluation further showcases TransDiffuser generates more diverse and plausible trajectories which explore more drivable area.
\end{abstract}

\section{Introduction}
Recently, substantial advancements have been attained across a diverse range of autonomous driving tasks including real-time localization \cite{vips,loc} and scene perception \cite{unitr,drivevlm}. 
Among them, planning-oriented autonomous driving \cite{drivevlm,uniad} has gained widespread attention from both academia and industry for its potential in improving traffic safety and efficiency. 
Early planning approach usually adopts a sequential process of perception, prediction, and subsequent planning, which causes information loss and cascading latency \cite{earlyPlanner}. 
Over the past few years, developing fully end-to-end planning-oriented autonomous driving systems \cite{uniad} has emerged as a key research direction.
Taking raw sensor data as input, an end-to-end driving model is expected to directly output an optimal trajectory for guiding future motion planning. 
Early works \cite{uniad,transfuser} aim to generate a single plausible trajectory in an auto-regressive manner by simply imitating annotated human expert driving behaviors in the training dataset.

However, this data-driven paradigm can be not fully capable to generalize to unseen and real-world scenarios.
Considering the existence of diverse yet complex driving scenarios and different feasible driving styles \cite{trajhf}, there is rarely  single feasible trajectory \cite{xing2025goalflow}. 
Recent attempts \cite{vad,hydramdp,xing2025goalflow} have increasingly focused on generating multi-mode trajectories as feasible candidates\footnote{In this paper, \textit{multi-mode trajectories} denote generating multiple possibly feasible candidate trajectories.}. 
To generate multi-mode trajectories from the continuous action space, one research line, represented by Hydra-MDP series \cite{hydramdp,hydramdp+,hydranext}, simplifies this challenge into selecting feasible candidates from a fixed planning vocabulary to discretize the action space. 
Another emerging research line \cite{xing2025goalflow,diffusiondrive,trajhf} aims to transfer the success of diffusion models \cite{ho2020denoising,dppo} to generate multi-mode trajectories, using scene and motion information as conditional input to produce multi-mode  trajectories as candidates.
GoalFlow \cite{xing2025goalflow} imposes a constraint on the trajectory generation process with scene priors, which is achieved by establishing a dense vocabulary of  goal points. 
DiffusionDrive \cite{diffusiondrive} further highlights the challenge of \textit{mode collapse}, wherein generated trajectories lack diversity as different noise inputs tend to converge to similar trajectory distribution after denoising.
It partitions the Gaussian distribution into multiple sub-Gaussian distributions centered around prior anchor trajectories for initialization. Notably, both previous research lines often require the pre-definition of trajectory vocabulary for anchor trajectories or pre-computation of scene priors. This introduces inductive bias which can face challenges for unseen scenarios. 

Following the intuition of the above Diffusion based approach, we propose TransDiffuser, an encoder-decoder based multi-mode trajectory generation model.
We utilize the frozen Transfuser backbone \cite{transfuser} to encode the scene perception information from front-viewed cameras and LiDAR.
The scene information and motion information of the current ego vehicle is then encoded as the conditional input of the Diffusion based denoising decoder. 
Unlike previous works, we identify another underlying bottleneck that leads to \textit{mode collapse} in generated trajectories: The under-utilization of the encoded multi-modal representation from  the conditional input of different modalities.
To explore the possibility of generating diverse and feasible trajectories without any anchors or scene priors, inspired by recent advance in self-supervised representation learning, we exploit a computation-efficient yet effective plug-and-play multi-modal representation decorrelation optimization mechanism during the denoising process, which aims to better exploit the multi-modal representation space to guide more diverse feasible planning trajectories from the continuous action space.
TransDiffuser achieves the new state-of-the-art performance on the planning-oriented NAVSIM benchmark without any explicit guidance like anchor-based trajectories or scene priors, and qualitative analysis further confirms it can generate diverse trajectories which explore more drivable space.
To sum up, our contributions are outlined as follows:

\begin{itemize}[leftmargin=0.2cm]
    \item We propose an encoder-decoder based generative trajectory model TransDiffuser. It firstly encodes the scene perception and motion state of the ego vehicle,
    and then utilizes the encoded information as conditional input of denoising decoder to decode multi-mode diverse yet feasible trajectories.
    \item Different from existing works that rely on predefined trajectories or pre-computed scene priors, we exploit a computation-efficient multi-modal representation decorrelation mechanism during the denoising process to enhance the diversity of generated trajectories to address the model collapse dilemma.
    \item We achieve the new state-of-the-art PDMS 94.85 on NAVSIM benchmark without any explicit guidance like predefined anchor trajectories or scene priors. Qualitative analysis showcases TransDiffuser generates diverse yet feasible trajectories which better explore the drivable space. 
\end{itemize}

\section{Related Works}
\label{sec:rw}

\subsection{End-to-end Autonomous Driving}
Modern autonomous driving systems \cite{uniad,diffusiondrive,hydramdp,trajhf} increasingly adopt end-to-end learning paradigms to directly map raw sensor data to driving decisions, streamlining the process from perception to action. Herein we review most recent related advances for planning and categorize current methods into three groups as shown in Table \ref{tab:review}.
\begin{figure}[htbp]
		\centering
		\includegraphics[width=0.8\linewidth]{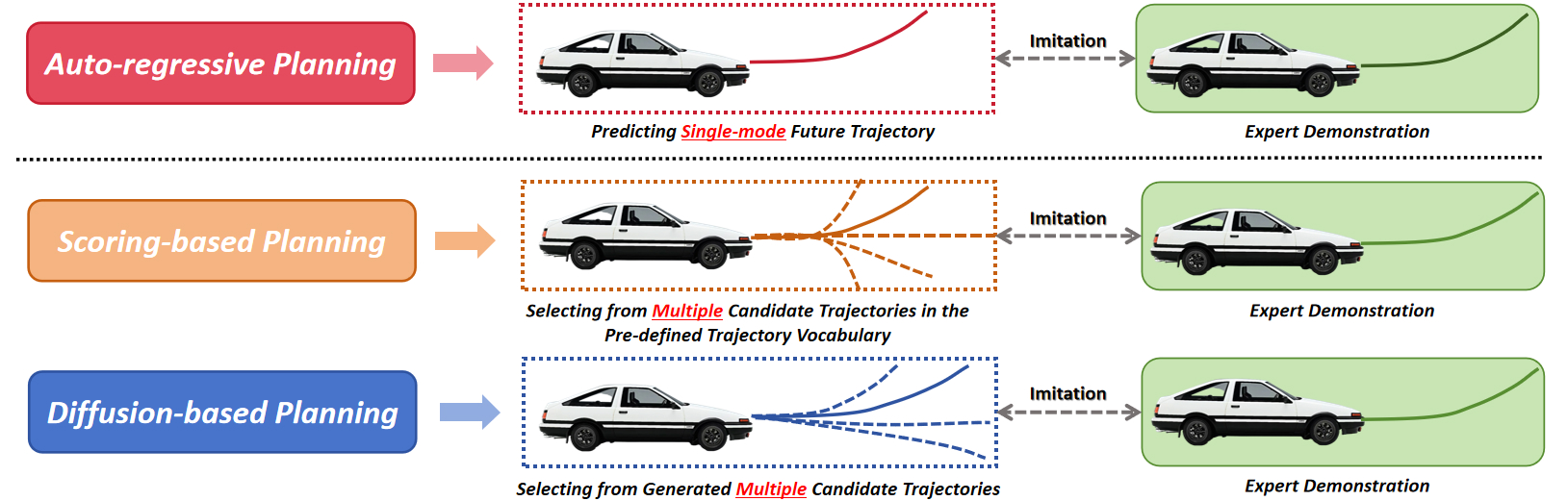}
		\caption{Main approaches for end-to-end autonomous driving.}
		\label{fig:vis_multi}
\end{figure}

\textbf{Auto-regressive Based Models.} This approach is relatively traditional, which typically predict a trajectory via sequential auto-regressive (\textbf{AR}) generating.
To our best knowledge, UniAD \cite{uniad} is the pioneering work showcasing the potential of end-to-end autonomous driving by integrating multiple perception tasks to facilitate planning. 
Transfuser \cite{transfuser} integrates image and LiDAR representations using self-attention and uses GRU to yield the planning trajectory in an auto-regressive paradigm. 
Its variant LTF is a light-weight image-only version where the LiDAR backbone is replaced by a learnable embedding.
PARA-Drive \cite{paradrive} proposes a computation-efficient system which performs mapping, planning, motion prediction and occupancy prediction tasks in parallel.

\textbf{Scoring Based Models.} This approach often generates multi-mode trajectories and selects optimal trajectory with the designed scoring functions.
VADv2 \cite{vadv2} is an early attempt to perform multi-mode planning by scoring and sampling from a large fixed vocabulary of anchor trajectories.
Centaur \cite{sima2025centaur} employs cluster entropy to measure uncertainty by analyzing the distribution of multiple trajectory candidates generated by the planner.
WoTE \cite{li2025end} leverages a latent BEV world model to forecast future BEV states for trajectory evaluation.
Hydra-MDP series \cite{hydramdp,hydramdp+,hydranext} convert the trajectory generation task into selecting optimal trajectory from predefined trajectory vocabulary. and propose an expert-guided hydra distillation strategy to align the planner with simulation-based metrics.
R2SE \cite{r2se} introduces a reinforced refinement framework with 3D backbone BEVFormer \cite{bevformer} that improves scaled hard case performance.

\textbf{Diffusion Based Models.}
One emerging trend aims to transfer the generative capabilities of difffusion models into the end-to-end planning task. Early attempts include GoalFlow \cite{xing2025goalflow}, DiffusionDrive \cite{diffusiondrive} and TrajHF \cite{trajhf}, which will be detailedly discussed in the later Section \ref{sec:genera}.

\begin{table}[htbp]
\caption{Comparison of  existing models. \textbf{*} denotes generating multi-mode candidate trajectories. }
\label{tab:review}
\centering
\resizebox{\linewidth}{!}{
\begin{tabular}{c|ccccc}
\toprule \toprule
\textbf{Model} & \textbf{Venue} & \textbf{Year} & \textbf{LiDAR} & \textbf{Anchor} & \textbf{Paradigm} \\ 
\midrule
Transfuser \cite{transfuser} & ICRA \& TPAMI & 2023 & $\checkmark$ & -- & AR \\
UniAD \cite{uniad} & CVPR & 2023 & -- & -- & AR \\
PARA-Drive \cite{paradrive} & CVPR & 2024 & -- & -- & AR \\
VADv2 \cite{vadv2} & -- & 2024 & -- & $\checkmark$ & Scoring*\\
Hydra-MDP \cite{hydramdp} & CVPRW & 2024 & $\checkmark$ & $\checkmark$ & Scoring*\\
DiffusionDrive \cite{diffusiondrive} & CVPR & 2025 & $\checkmark$ & $\checkmark$ & Diffusion*\\
GoalFlow \cite{xing2025goalflow} & CVPR & 2025 & $\checkmark$ & $\checkmark$ & Diffusion*\\
Hydra-MDP++ \cite{hydramdp+} & -- & 2025 & -- & $\checkmark$ & Scoring*\\
Hydra-NeXt \cite{hydranext} & -- & 2025 & -- & $\checkmark$ & Scoring*\\
R2SE \cite{r2se} & -- & 2025 & -- & -- & Scoring* \\
TrajHF \cite{trajhf} & -- & 2025 & $\checkmark$ & -- & Diffusion* \\
DIVER \cite{breaking} & -- & 2025 & $\checkmark$ & -- & Diffusion* \\
Centaur \cite{sima2025centaur} & -- & 2025 & $\checkmark$ & $\checkmark$ & Scoring* \\
WoTE \cite{li2025end} & ICLR & 2025 & $\checkmark$ & $\checkmark$ & Scoring* \\
\midrule
\textbf{TransDiffuser} & -- & 2025 & $\checkmark$ & -- & Diffusion*\\
\bottomrule
\end{tabular}
}
\end{table}

\subsection{Generative Trajectory Model}
\label{sec:genera}

Generative models have garnered significant attention due to their remarkable capabilities in producing realistic data across various domains.
These models have been successfully applied in numerous tasks, such as image synthesis using Generative Adversarial Network \cite{gan} and Variational Autoencoder \cite{vae} and text generation with large language models \cite{gpt4}.
Diffusion models have been proposed and widely studied in recent years, and demonstrated their powerful generative abilities across vision generation \cite{ho2020denoising,fd,dip,laf} and language generation \cite{llada}.
A recent shift has emerged in the field of end-to-end autonomous driving, where researchers \cite{diffusiondrive,xing2025goalflow,trajhf} embark on exploring Diffusion models for planning. 
Diffusion models, known for their ability to generate high-quality data by iteratively denoising random noise, have shown promise in generating smooth and realistic driving trajectories. 
However, pointed by DiffusionDrive \cite{diffusiondrive}, one bottleneck lies in that diffusion based trajectory generative models showcases the less diversity of the decoded denoised trajectories. \textit{This issue is often called by mode collapse}. 
It proposes truncated diffusion policy that begins the denoising process from an anchored gaussian distribution instead of a standard Gaussian distribution. 
GoalFlow \cite{xing2025goalflow} used a goal point vocabulary to mitigate the \textit{mode collapse} issue to assist the efficient trajectory generation.
TrajHF \cite{trajhf} pioneers to introduce preference optimization on their private dataset to align the generative trajectory model.
Different from above works, we aim to mitigate the \textit{mode collapse} issue via improving the fused intermediate multi-modal representations, which assists to increase the diversity of final decoded trajectories. Notably, we do not utilize any form of guidance like anchors or scene priors, allowing for more flexible and diverse planning.

\subsection{Representation Learning}

Representation learning has shown its potential in many deep learning applications like self-supervised learning \cite{fdssl}, supervised classification \cite{decorr,fedlf,sasvd}, noisy label learning \cite{fnbench,yihao,dualoptim} and reinforcement learning \cite{fdrl}. In general, these attempts aim to fully exploit the representation space to learn better intermediate representations for downstream applications. One popular approach is contrastive learning  \cite{simclr} by mining the representation similarities among positive and negative examples. 
However, this approach generally requires very large training batch (e.g. 4096 or 8192) to fully extract reliable supervision from batch examples, which may be not suitable and computation-efficient for training autonomous driving models. 
Inspired by the recent advance \cite{fdrl,fnbench,decorr}, we optimize the intermediate representations in our task by decoupling the interactions of different representation dimensions. This optimization approach does not rely on the large training batchsize and bring little computation overhead as analyzed in \cite{fdrl,decorr}, which we will introduce in Section \ref{sec:fd} and \ref{sec:sensi}.


\section{Method}

\subsection{Preliminary}
\begin{figure*}[htbp]
		\centering
		\includegraphics[width=0.85\textwidth]{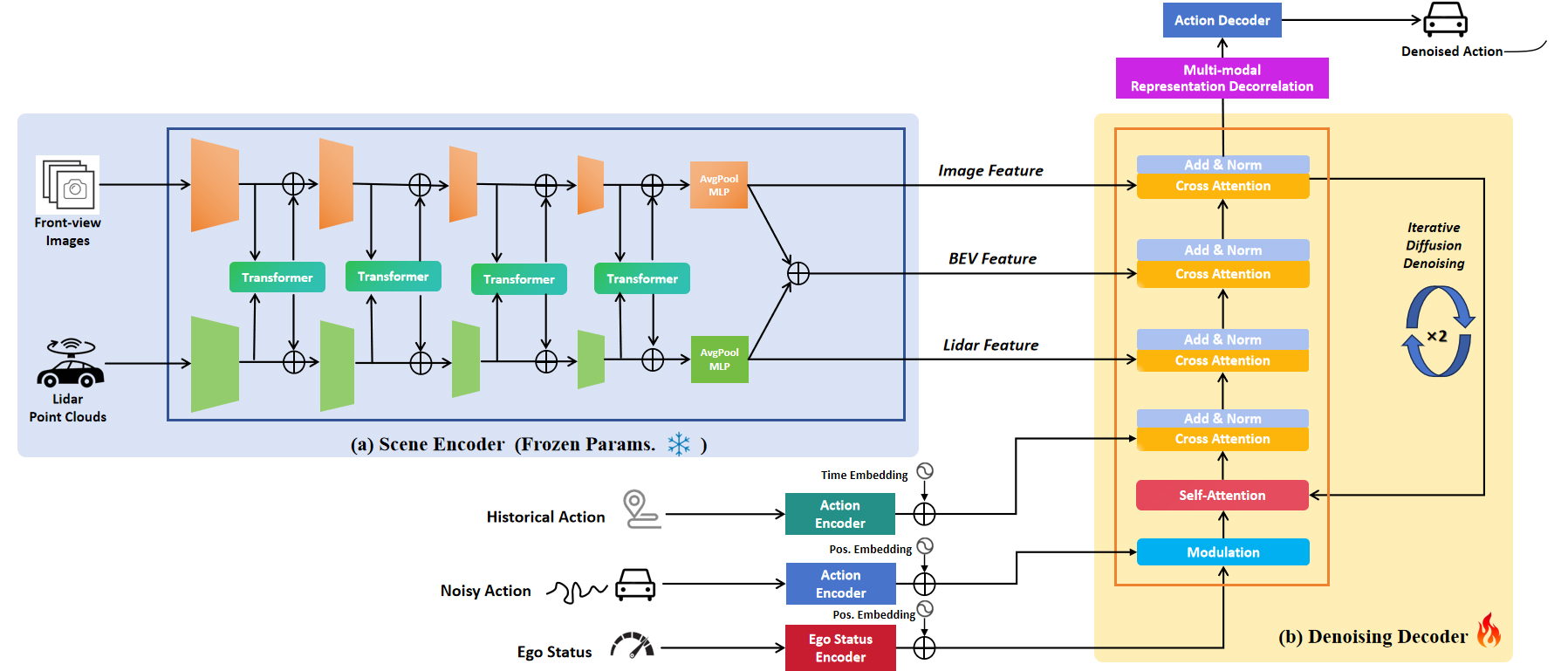}
		\caption{Overview of the proposed TransDiffuser architecture. We freeze the parameters of the scene encoder. 
        }
		\label{fig:framework}
\end{figure*}

As shown in Fig. \ref{fig:framework}, our framework is generally an encoder-decoder model, which consists of two main components: Scene encoder and denoising decoder.
It takes raw sensor data as input and predicts the future trajectory by accumulating consequent decoded actions.
The derived trajectory is represented as a sequence of waypoints (states) $x=\{s_1,s_2,\ldots,s_\mathcal{T}\}$, where where $\mathcal{T}$ denotes the trajectory length, and each state $s_{\tau}$  is the location of the $\tau-th$ waypoint in the current driving agent's ego-centric coordinate system.
Following \cite{trajhf}, each waypoint can connect to its neighbor waypoints by sequentially projecting to action space to mitigate heteroscedasticity along the trajectory timeline. The projection can be recursively expressed as:
\begin{equation}
\hat{x}_\tau, \hat{x}_1=s_\tau-s_{\tau-1},s_1
\end{equation}
where $\hat{x}_\tau$ represents the agent's action at timestep $\tau$, and $\tau$ ranges from 2 to $\mathcal{T}$.
This mapping guarantees a reversible connection between the two spaces. As a result, the trajectory can be easily deduced through the accumulation of actions.

\subsection{Scene \& Motion Encoder}
We utilize a Scene Encoder (Figure \ref{fig:framework}(a)) based on the Transfuser backbone \cite{transfuser}, pre-trained on Nav-train, to process rich perception information from front-view cameras and LiDAR sensors. This multi-modal approach is crucial as single modalities often lack essential environmental details. Image and LiDAR information are processed through separate backbone networks (dependent branches). We perform multi-stage fusion by connecting features from corresponding stages of both branches via Transformer blocks \cite{transformer}. This allows for cross-modal attention, leading to enhanced multi-scale representations and better overall integration of sensor data.
The final output of Scene Encoder include the image feature $F_{img}$, LiDAR feature $F_{LiDAR}$ and BEV feature $F_{bev}$. 

To provide essential motion context for predicting future actions, the encoder processes two types of motion information. Firstly, the historical ego trajectory is encoded into an action embedding $Emb_{action}$ using a dedicated Multi-Layer Perceptron (MLP), referred to as the Action Encoder. Secondly, the current ego vehicle status is encoded into an ego status embedding $Emb_{ego}$ by another MLP (Ego Status Encoder). Following \cite{trajhf}, these two motion-related embeddings are designed to be utilized within the denoising decoder.
The encoded scene and motion feature group $feat=\{F_{bev}, F_{img}, F_{LiDAR}, Emb_{action}. Emb_{ego}\}$ is the conditional input of our denoising decoder.

\subsection{Denoising Decoder}
\label{sec:denoiser}

The denoising decoder (Figure \ref{fig:framework}(b)) is responsible for generating the planned trajectory by iteratively refining an initial noise estimate, conditioned on the encoded scene and motion features derived from the Scene Encoder.
It takes the encoded feature group $feat = \{F_{bev}, F_{img}, F_{LiDAR}, Emb_{action}, Emb_{ego}\}$ as conditional input. Different features within this group are sequentially fused via multi-head cross-attention~\cite{transformer}. The decoder then outputs a feasible action from the continuous action space. By periodically accumulating these decoded actions, the final trajectory is formed.
To achieve this generation and handle the complexity of driving scenarios, we employ Denoising Diffusion Probabilistic Models (DDPM) \cite{ho2020denoising} as the optimization framework. DDPM involves forward process where noise is gradually added to the ground truth data, and reverse process where the model learns to denoise.
Following~\cite{trajhf,diffusiondrive}, we focus on reverse denoising process, which transitions from Gaussian noise towards noise-free state $x_0$. This transition is governed via this equation:
\begin{equation} \label{eq:denoising_step}
x_{t-1} = \frac{1}{\sqrt{\alpha_t}}(x_t - \frac{1-\alpha_t}{\sqrt{1-\bar{\alpha}_t}}\epsilon_{\theta}(x_t, t, feat)) + \sigma_t z
\end{equation}
where $z \sim \mathcal{N}(0, I)$, $t \in \{1, ..., \mathbf{T}\}$ denotes the noise level timestep, $\mathbf{T}$ is the total number of denoising steps, and $\epsilon_{\theta}$ represents our denoising decoder parameterized by $\theta$. The noise prediction $\epsilon_{\theta}$ is conditioned on the noisy state $x_t$, the timestep $t$, and the feature group $feat$. The parameters $\alpha_t = 1 - \beta_t$, $\bar{\alpha}_t = \prod_{i=1}^{t} \alpha_i$, and $\sigma_t^2 = \beta_t$ are derived from a predefined noise scheduler $\beta_t$. Eq.~\ref{eq:denoising_step} describes how to estimate a slightly less noisy state ($x_{t-1}$) from the current noisy state ($x_t$) using the model's noise prediction ($\epsilon_{\theta}$).
The decoder is trained to predict the noise added during the forward process. The optimization objective involves minimizing the difference between the actual sampled noise $\epsilon$ and predicted noise $\epsilon_{\theta}$, formulated as a gradient descent step on:
\begin{equation} \label{eq:loss}
\nabla_{\theta} ||\epsilon - \epsilon_{\theta}(\sqrt{\bar{\alpha}_t}x_0 + \sqrt{1-\bar{\alpha}_t}\epsilon, t, feat)||^2
\end{equation}
where $\epsilon$ is randomly sampled from $\mathcal{N}(0, I)$. Detailed computation of trajectory denoising loss, $\mathcal{L}_{diff}$, using Mean Squared Error (MSE) is outlined in Algorithm~\ref{algo:traj_loss}.
In detail, TransDiffuser denoises an 8-step action sequence in parallel during inference which is then cumulatively summed to recover the multiple key waypoints of the complete trajectory.

\begin{algorithm}[htbp]
\caption{The computation process of trajectory denoising loss $\mathcal{L}_{diff}$}
\label{algo:traj_loss}
\LinesNumbered
\KwIn{Trajectory GT $label$, Scene feature $feat$, Number of denoising timesteps $\mathbf{T}$}
\KwOut{Trajectory loss $\mathcal{L}_{diff}$}
\textcolor{blue}{// Generate random Gaussian noise} ; \\
$noise \gets \text{Gaussian}(label.shape)$ ; \\
\textcolor{blue}{// Pick a random time step} ; \\
$ step \gets \text{Rand}(0, \mathbf{T})$ ; \\
\textcolor{blue}{// Apply noise in the forward process} \\
$noisy\_target \gets \text{AddNoise}(label, noise, step)$ ; \\
\textcolor{blue}{// Denoising process} \\
$pred \gets \text{Denoising\_Decoder}(noisy\_target, step, feat)$ ; \\
$\mathcal{L}_{diff} \gets \text{MSE}(pred, noise)$ ; \\ 
\Return $\mathcal{L}_{diff}$ ;
\end{algorithm}

During the inference process, the model leverages this learned denoising process to generate a pool of diverse candidate trajectories starting from pure Gaussian noise. In our implementation, we generate $N=30$ candidate trajectories considering the efficiency. Subsequently, rejection sampling strategy is employed to filter out dynamically infeasible or unsuitable trajectories following TrajHF \cite{trajhf}. sensitivity analysis on the candidate number is provided in Section \ref{sec:sensi}.

\textbf{\textit{Discussion:}} GoalFlow \cite{xing2025goalflow} typically generates 128 or 256 candidate trajectories, while TrajHF \cite{trajhf} uses 100 candidates prior to their respective selection or filtering steps. 
As analyzed in Section \ref{sec:explore}, we generate fewer candidates but still maintain effective trajectory quality, our method naturally embodies a degree of inference efficiency.

\subsection{Multi-modal Representation Decorrelation}
\label{sec:fd}

\begin{figure}[htbp]
		\centering
		\includegraphics[width=1\columnwidth]{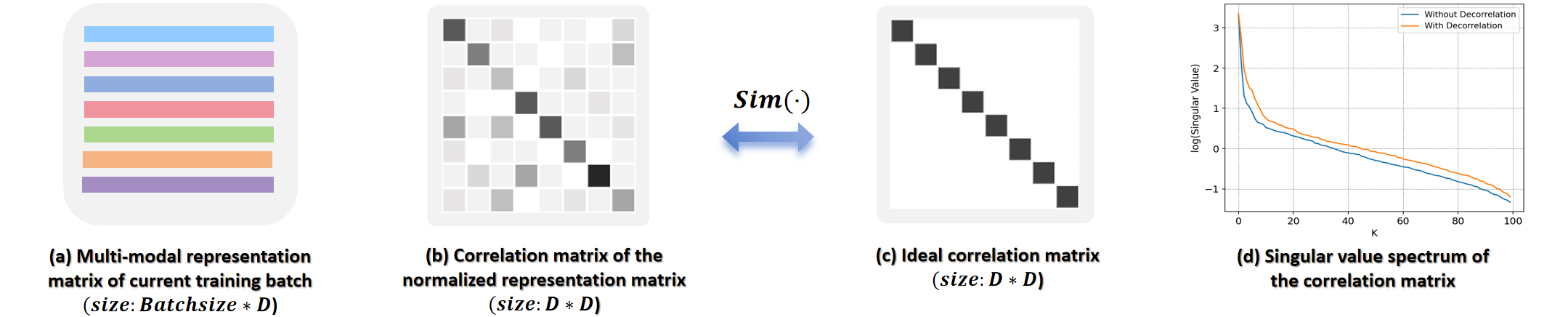}
		\caption{Illustrations on the multi-modal representation decorrelation process. (d) shows top 100 largest singular values of the correlation matrix. We observe the decorrelation mechanism helps to better utilize the representation space before action decoding (shown in Figure \ref{fig:framework}).}
		\label{fig:MFD}
\end{figure}

The final denoised action is decoded by the action decoder as illustrated in Figure \ref{fig:MFD}. 
We argue planning performance is determined by the quality of the learned multi-modal representations to certain degree.
To tackle this information bottleneck \cite{bottleneck}, we apply the multi-modal representation optimization objective on the fused multi-modal representations, which is illustrated in Figure \ref{fig:MFD}.
In each training batch $B$, we first normalize multi-modal representation matrix $\mathbf{M}$ and then compute its correlation matrix $\textbf{corr}$.
This penalty $\mathcal{L}_{reg}$ is designed to decrease non-diagonal entries of the multi-modal correlation matrix $\textbf{corr}$.

Intuitively, $\mathcal{L}_{reg}$ aims to regularize the multi-modal correlation matrix (i.e. Figure \ref{fig:MFD}) to be similar to the diagonal form. This can be achieved by eliminating the unnecessary interactions of different dimensions of representation to increase the similarity (illustrated as $sim(\cdot)$) to identity matrix or diagonal matrix (i.e. Figure \ref{fig:MFD}(c)). 
As the figure shows, the derived $\mathcal{L}_{rep}$ is a batch-level regularization objective which optimizes the multi-modal representations in each training batch.
Since the singular values of the covariance matrix \textbf{corr} provide a comprehensive characterization \cite{decorr} of the distribution of multi-modal representations $\textbf{M}$, we visualize the largest singular values shown in Figure \ref{fig:MFD}(d).
We also provide the batch-level sensitivity study in Section \ref{sec:sensi}.

\begin{algorithm}[htbp] 
    \caption{The computation process of $L_{reg}$}
    \label{algo:decorrloss}
    \LinesNumbered
    \KwIn{Reshaped multi-modal representation $\mathbf{M}$}
    \KwOut{$L_{reg}$}
    $\mathbf{M}, \sigma_{\mathbf{M}} \gets \mathbf{M} - \text{Mean}(\mathbf{M},  \text{keepdim}=\text{True}), \text{Var}(\mathbf{M},  \text{keepdim}=\text{True})$ \;
    \textcolor{blue}{//Normalization. $\epsilon=1e^{-8}$ is used to avoid zero denominator.}  \\
    $\mathbf{M} \gets \frac{\mathbf{M}}{\sqrt{\epsilon + \sigma_{\mathbf{M}}}}$ ;  \\
    \textcolor{blue}{//Compute the correlation matrix.}  \\ 
    $\mathbf{corr} \gets \mathbf{M}^T \cdot \mathbf{M}$ ;  \\
    \textcolor{blue}{//Extract the non-diagonal elements.}  \\
    $\hat{corr} \gets \text{remove\_diagonal\_elements} (\mathbf{corr})$ ;  \\ 
     $\mathcal{L}_{rep} \gets (\hat{corr}^2.mean())/|\mathbf{B}|$  \;
    \Return $\mathcal{L}_{rep}$ \;
\end{algorithm}

We combine the supervised $\mathcal{L}_{diff}$ and the multi-modal representation decorrelation loss with the trade-off coefficient $\beta$ during training:
\begin{equation}
\label{eq:all}
    \mathcal{L} = \mathcal{L}_{diff} +\beta\mathcal{L}_{rep}.
\end{equation}

\textbf{\textit{Discussion:}} Herein we simply clarify this novelty. Though representation optimization is relatively widely studied in traditional machine learning tasks \cite{yuandong,fnbench,sasvd}, it is rarely considered in generative models \cite{diffuse}, especially for trajectory generation oriented at end-to-end autonomous driving. In the meantime, decorrelation related computation overhead (FLOPs) is  less than 1‰ of the forward process and it benefits the generation diversity as shown in Figure \ref{fig:vis_multi}.


\subsection{Quantification of Mode Collapse}
\label{sec:quanti}
To evaluate the multi-mode property of the generative trajectory models, we follow DiffusionDrive \cite{diffusiondrive} to develop the quantification analysis.
As analyzed in previous diffusion based studies \cite{diffusiondrive,xing2025goalflow}, different initialized random noises often lead to converged similar trajectories after the denoising process.
To quantitatively analyze the phenomenon of \textit{mode collapse}, we formulate the mode diversity score $\mathcal{D}$ based on the mean Intersection over Union (mIoU) between each denoised trajectory and the union of all denoised trajectories:

\begin{equation}
    \mathcal{D} = 1 - \frac{1}{N} \sum_{i=1}^{N} \frac{\text{Area}\left(\tau_{i} \cap \bigcup_{j=1}^{N} \tau_{j}\right)}{\text{Area}\left(\tau_{i} \cup \bigcup_{j=1}^{N} \tau_{j}\right)}
\end{equation}
where $\tau_i$ represents the $i-th$ denoised trajectory, $N$ is the total number of sampled trajectories and  $\tau_j$ is the union of all denoised trajectories, as introduced in Section \ref{sec:denoiser}.

\section{Experiments}
\label{sec:exp}

\subsection{Experimental Setup}

\textbf{Dataset.} 
Following previous state-of-the-art works \cite{diffusiondrive,xing2025goalflow,trajhf}, We utilize the well-established planning-oriented NAVSIM dataset \cite{dauner2024navsim} using non-reactive simulation and closed-loop evaluations.
It builds on the existing OpenScene \cite{openscene} dataset, a compact version of nuPlan \cite{nuplan} sampled at 2 Hz. 
Each sample contains camera images from 8 perspectives, fused LiDAR data from 5 sensors, ego status, and annotations for the map and objects.
It has two parts: Nav-train and Nav-test, which respectively contain 1192 and 136 scenarios for training/validation and testing, and we only use the training split for training and validation set to guide model selection.

\textbf{Baseline.} 
We compare the proposed framework against counterparts 
from three groups discussed in Section \ref{sec:rw}.
Following ~\cite{dauner2024navsim,xing2025goalflow}, we additionally refer to  \textit{Constant Velocity} and \textit{Ego Status MLP} as the lower bound for comparison. 
\textit{Constant Velocity} model assumes constant speed from the current timestamp for forward movement. 
\textit{Ego Status MLP} serves as a bind driving agent, which leverages a MLP for trajectory prediction given only the ego vehicle status.

\begin{table*}[htbp]
    \caption{Performance on NAVSIM. \textbf{L}, \textbf{V} and \textbf{V*}  denote LiDAR input, vision input, and video or historical image input. }
    \label{tab:main}
    \centering
    \begin{tabular}{c|cc|cccccc}
        \toprule
        \toprule
\multirow{2}{*}{\textbf{Method}} & \multirow{2}{*}{\textbf{Modality}} & \multirow{2}{*}{\textbf{Image Encoder}} & \multicolumn{6}{c}{\textbf{Metrics}} \\
 &  & & $NC\uparrow$ & $DAC\uparrow$ & $EP\uparrow$ & $TTC\uparrow$ & $Comfort\uparrow$& $\textbf{PDMS}\uparrow$ \\ \midrule
 Constant Velocity \cite{dauner2024navsim} & - & -& 68.0& 57.8& 19.4& 50.0& 100&20.6\\
 Ego Status MLP \cite{dauner2024navsim} & -& -& 93.0& 77.3& 62.8& 83.6& 100&65.6\\ \midrule
        Transfuser ~\cite{transfuser} &V+L& ResNet-34&97.8 & 92.6 & 78.9 & 92.9  & 100& 83.9 \\
 LTF \cite{transfuser} & V& ResNet-34& 97.4& 92.8& 79.0& 92.4& 100&83.8\\
 UniAD \cite{uniad} & V+L& ResNet-34& 97.8& 91.9& 78.8& 92.9& 100&83.4\\
 PARA-Drive \cite{paradrive}& V*+L& ResNet-34& 97.9& 92.4& 79.3& 93.0& 99.8&84.0\\ \midrule
        DiffusionDrive ~\cite{diffusiondrive}   &V+L& ResNet-34&98.2 & 96.2 & 82.2 & 94.7  & 100& 88.1 \\
        GoalFlow ~\cite{xing2025goalflow} &V+L& ResNet-34&98.4 & 98.3 & 85.0 & 94.6  & 100& 90.3  \\
        VADv2 \cite{vadv2}& V+L& ResNet-34& 97.2& 89.1& 76.0& 91.6& 100&80.9\\
        Hydra-MDP ~\cite{hydramdp}  &V+L& ResNet-34&99.1 & 98.3 & 85.2 & 96.6  & 100& 91.3 \\ 
        Hydra-NeXt ~\cite{hydranext} & V & ResNet-34&  98.1 & 97.7 & 81.8 & 94.6 & 100 & 88.6 \\
        Hydra-MDP++ (Base) ~\cite{hydramdp+}  &V*& ResNet-34& 97.6& 96.0& 80.4& 93.1& 100& 86.6\\
        WoTE \cite{li2025end} & V+L& ResNet-34&98.5 & 95.8 & 80.9 & 94.4  & 99.9 & 87.1 \\
        Hydra-MDP++ (Large) ~\cite{hydramdp+}  &V*& V2-99 & 98.6& 98.6& 85.7& 95.1& 100& 91.0\\ 
        R2SE \cite{r2se} &V & BEVFormer & 99.0 & 97.9 & 86.8 & 96.4 & 100  & 91.6 \\ 
        Centaur \cite{sima2025centaur} & V+L &ResNet-34&\textbf{99.5} & \textbf{98.9} & 85.9 & \textbf{98.0}  & 100 & 92.6 \\ 
        TrajHF \cite{trajhf} & V+L& ViT&99.3 & 97.5 & 90.4 & \textbf{98.0}  & 99.8 & 94.0 \\  
        DIVER \cite{breaking} & V+L &ResNet-34& 98.5 & 96.5 & 82.6 & 94.9  & 100 & 88.3 \\ 
        \midrule
        \textbf{TransDiffuser (Ours)}& V+L & ResNet-34& 99.4 & 96.5 &\textbf{94.1} & 97.8 & 99.4 & \textbf{94.9}\\
        \bottomrule 
    \end{tabular}
\end{table*}

\textbf{Implementation Details.}
Our code is built on the PyTorch lightening framework \cite{pytorchlightening} and official NAVSIM toolkit \cite{dauner2024navsim}.
We take 10 timesteps during denoising  process for both training and inference. The trade-off coefficient $\beta$ in Eq. \ref{eq:all} is fixed to $0.02$.
The learning rate is set to $1e-4$, and global batchsize is $256$, distributed across $4$ NVIDIA$^\circledR$ H20 GPUs. 
Adam optimizer \cite{adam} is adopted with OneCycle scheduler. 
As the task requires, the model outputs $8$ waypoints spanning $4$ seconds. 
Total training lasts for $120$ epochs. 

\textbf{Metrics.} For NAVSIM dataset, we evaluate our models based on the popular close-looped Predictive Driver Model score (PDMS \cite{pdms}) which quantifies driving capacity by aggregating sub-metrics for multiple objectives such as progress and comfort. It can be formulated as follows:
\begin{align}
PDMS &= NC \times DAC \times TTC \times \notag \\
&\quad \frac{(5 \times DDC + 2 \times C + 5 \times EP)}{12},
\end{align}
where the sub-metrics $NC$ (No at-fault Collision), $DAC$
(Drivable Area Compliance), $EP$ (Ego Progress), $Comfort$, $DDC$ (Driving Direction Compliance), and $TTC$ (Time-to-Collision), each represented as a percentage, are composed into a single score \cite{dauner2024navsim}. $DDC$ is exempted from calculation due to the practical constraints of the NAVSIM toolkit\footnote{\href{https://github.com/autonomousvision/navsim/issues/14}{https://github.com/autonomousvision/navsim/issues/14}}. 
Following \cite{diffusiondrive,trajhf}, the top-1 scoring predicted trajectory for each sample is used for evaluation.
To measure the diversity of all generated candidate trajectories for each sample, we refer to \cite{diffusiondrive} for \textit{Diversity} metric $\mathcal{D}$.

\subsection{Main Experiments}
\label{sec:mainexp}
\begin{figure}[htbp]
		\centering
		\includegraphics[width=\linewidth]{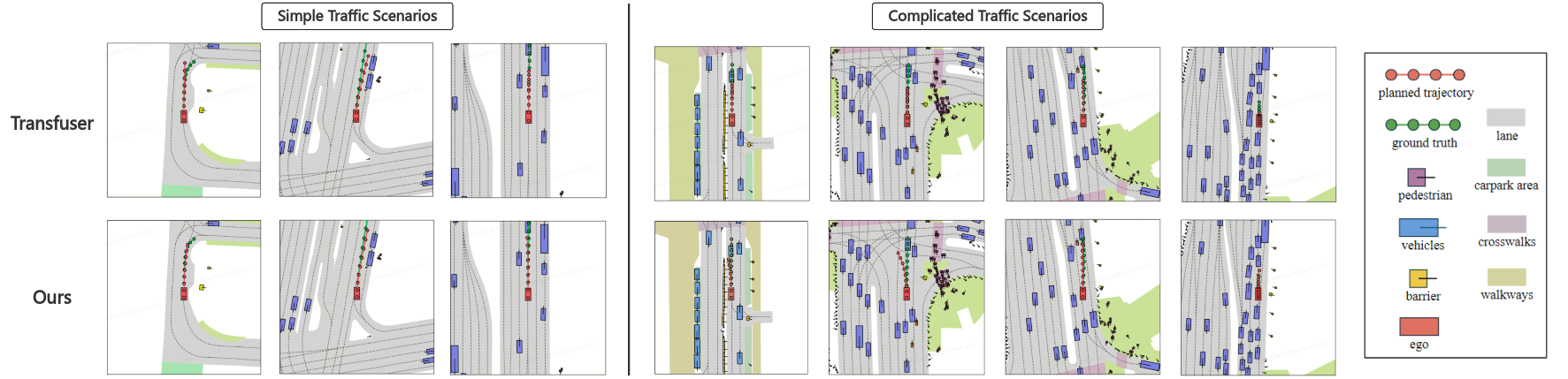}
		\caption{Visualization of single-mode trajectories.}
		\label{fig:vis}
\end{figure}

\textbf{Benchmark Performance.}
As Table \ref{tab:main} indicates, TransDiffuser achieves the best performance on Nav-test against baseline methods like DiffusionDrive and GoalFlow. Performance of other models are cited from the original papers. Hydra-MDP++ provides a base version and a large version by scaling up the image encoder from ResNet-34 \cite{resnet} to V2-99 \cite{v2-99}. 
The most obvious improvement of TransDiffuser lies in Ego Progress metric, which can also be intuitively observed in Figure \ref{fig:vis} and \ref{fig:vis_multi}. We also witness the potential of Diffusion based models. Note that different from previous state-of-the-art methods like GoalFlow and DiffusionDrive, we do not rely on any priors as discussed in Section \ref{sec:rw}.

%

\textbf{Diversity Improvement.} Following the diversity metric in Section \ref{sec:quanti}, as shown in Table \ref{tab:abla} and \ref{tab:expl}, we measure the $\mathcal{D}iversity$ metric over the test dataset. 
We observe the existence of multi-modal representation decorrelation regularization correspondingly improves the diversity of generated candidate trajectories (from $66$ to $70$) while there still exists a small gap to DiffusionDrive ($74$) which benefits from additional trajectory vocabulary to initialize the anchored distribution.
In general, TransDiffuser can yield relatively diverse trajectories without any additional priors like anchored trajectories or scene priors.
We conduct further analysis on how denoising steps and candidate number affect the diversity in Section \ref{sec:sensi} and \ref{sec:explore}, which shows that TransDiffuser can achieves better diversity with more denoising timesteps, and it achieves better performance with 10/15 candidates than 20 candidates of DiffusionDrive.

\textbf{Training Efficiency.} Our training performs 120 epochs across four GPUs and consumes within 2 wall-clock hours. TransDiffuser model is about 251M parameters while the parameters of the perception encoder are frozen, thus only 62.8M parameters are trainable. Regarding inference-time efficiency, we discussed in Section \ref{sec:denoiser}.

\subsection{Sensitivity Study}
\label{sec:sensi}
Herein we study three key hyper-parameters: Diffusion timesteps $\mathbf{T}$ in Section \ref{sec:denoiser}, Batchsize $\mathbf{B}$ in Section \ref{sec:fd} and $\beta$ in Eq.\ref{eq:loss}. As Table \ref{tab:abla} indicates, our method shows robustness over the selection of coefficient $\beta$, diffusion timesteps $\mathbf{T}$ and batchsize $\mathbf{B}$. We obverse when we scale up the diffusion timesteps, the diversity metric $\mathcal{D}$ generally increases at the cost of more computation and latency, while small diffusion timesteps can also achieve satisfactory performance with higher efficiency and relatively lower diversity. Future works can consider more advanced policies like flow matching. 

\vspace{-2pt}
\begin{table}[htbp]
    \caption{Experimental results for the sensitivity study. }
    \label{tab:abla}
    \centering
    \small
    \resizebox{\linewidth}{!}{
\begin{tabular}{cc|ccccccc}
\toprule \toprule
\multirow{2}{*}{\textbf{Component}}& \multirow{2}{*}{\textbf{Value}}& \multicolumn{7}{c}{\textbf{Metrics}} \\
 &  & $NC$ & $DAC$ & $EP$ & $TTC$ & $Comfort$ & \textbf{PDMS} & \textbf{$\mathcal{D}$} \\ \midrule
\multirow{3}{*}{Timestep $\mathbf{T}$} & 5 &  98.5&  94.4&  92.7&  95.8&  99.8& 92.4& 65 \\
 & 10 &  99.4&  96.5&  94.1&  97.8&  99.4&  94.9 & 70 \\
 & 20 &  99.0&  96.0&  94.6&  96.7&  99.3&  94.3& 88 \\ \midrule
\multirow{3}{*}{Batchsize $\mathbf{B}$} & 32 &  98.8&  94.6&  92.9&  96.1&  99.9&  92.7 & 66 \\
 & 64 &  99.4&  96.5&  94.1&  97.8&  99.4&  94.9 & 70 \\
 & 128 &  98.9&  95.4&  91.7&  97.0&  99.0&  92.9 & 69 \\ \midrule
\multirow{4}{*}{Coefficient $\beta$} & 0 &  99.0&  95.8&  94.5&  96.8&  99.9&  94.3 & 66 \\
 & 0.02 &  99.4&  96.5&  94.1&  97.8&  99.4&  94.9 & 70 \\
 & 0.05 &  99.0&  95.9&  93.8&  97.2&  99.3&  94.1 & 69 \\
 & 0.1 &  99.0&  95.9&  93.3&  97.3&  99.1&  94.0 & 69 \\ \bottomrule
\end{tabular}
}
\end{table}

\subsection{Exploration Experiments}
\label{sec:explore}
Herein we discuss some key exploration regarding our TransDiffuser and multi-modal decorrelation optimization. Experimental results are shown in Table \ref{tab:expl}.

\textbf{Candidiate number.} The number of candidate trajectories is a key factor to performance. Previous methods like GoalFlow and TrajHF require no less than 100 candidate trajectories. 
With the assistance of anchored trajectory distribution and optimized generation policy, DiffusionDrive \cite{diffusiondrive}  exhibits satisfactory performance by generating 20 candidates. We decrease the candidate number to 10 and 15, our proposed TransDiffuser still shows its robustness on planning while fewer candidates yield the inferior diversity. 

\textbf{Inner or outer decorrelation.} In our original design in Figure \ref{fig:framework}, the decorrelation is adopted on the outside of denoising decoder and before the action decoder. We also explore when we apply this decorrelation mechanism inside the denoising loop within the denoising decoder, and the results indicate this inner decorrelation shows slight performance and diversity decrease.

\textbf{Holistic training.} Scene encoder is frozen in TransDiffuser since we observe the slight performance decrease with the consistent diversity when we fully train the holistic model. 

\textbf{Generalization potential.} We apply this decorrelation optimization on the auto-regressive based Transfuser, we find the planning performance gets improved. 

%

\begin{table}[htbp]
\centering 
\caption{Performance on the exploration experiments.}
\label{tab:expl}
\resizebox{\columnwidth}{!}{ 
\begin{tabular}{c|ccc|c|c}
\toprule \toprule
\textbf{Method} & \textbf{Img. Encoder} & \textbf{Anchor} & \textbf{Candidate(s)} & \textbf{PDMS} & \textbf{$\mathcal{D}$} \\ \midrule
Transfuser ($\dagger$)& ResNet-34 & $\times$ & 1 & 78.0 & -\\
Transfuser ($\dagger$) + Decorr.& ResNet-34 & $\times$ & 1 & 78.8 & -\\ \midrule
GoalFlow & ResNet-34 & $\checkmark$ &  128/256 & 90.3 & N/A \\
DiffusionDrive & ResNet-34 & $\checkmark$ & 20 & 88.1 & 74 \\
TrajHF & ViT & $\times$ & 100 & 94.0 & N/A \\ \midrule
Ours & ResNet-34 & $\times$ & 10 & 89.6 & 56 \\
Ours & ResNet-34 & $\times$ & 15 & 91.7 & 63 \\
Ours & ResNet-34 & $\times$ & 30 & 94.9 & 70 \\ \midrule
Ours (Fully-trained) & ResNet-34 & $\times$ & 10 & 88.7 & 55\\
Ours (Fully-trained) & ResNet-34 & $\times$ & 15 & 90.8 & 60 \\
Ours (Fully-trained) & ResNet-34 & $\times$ & 30 & 93.5 & 68\\ \midrule
Ours (Inner.) & ResNet-34 & $\times$ & 10 & 86.8 & 56 \\
Ours (Inner.) & ResNet-34& $\times$ & 15 & 89.8 & 63\\
Ours (Inner.) & ResNet-34& $\times$ & 30 & 93.5 & 69 \\ \bottomrule
\end{tabular}
}
\end{table}
\vspace{-2pt}

\subsection{Qualitative Analysis}
\textbf{Comparison with Transfuser.}
Since TransDiffuser endows the auto-regressive based Transfuser \cite{transfuser} with generation capabilities to some extent, we provide the visualization of representative examples from the BEV perspective. 
We visualize the single auto-regressive trajectory predicted by Transfuser and the single selected trajectory of our model.
For simple traffic scenarios where there are fewer notable objects, TransDiffuser can propose relatively more aggressive feasible planning trajectories.
For complicated traffic scenarios where there are numerous notable objects around the ego driving agent, it tends to propose diverse yet feasible planning trajectories on the premise of ensuring safety.

\textbf{Comparison with GoalFlow.}
Since diffusion based counterparts like GoalFlow \cite{xing2025goalflow} do not provide specific scene tokens for visualization, we take similar scenes as examples to compare the diversity of generated trajectories in Figure \ref{fig:vis_multi}. 
Though we sampled 30 trajectories (fewer than 128/256 trajectories in GoalFlow), we can get more diverse yet feasible trajectories covering more plausible driving space. 
In contrast, trajectories generated by GoalFlow converge to the similar trajectory distribution given the guidance of goal point priors.
By exploiting decorrelation optimization to enrich latent representation space, we can explore more diverse  trajectories.

\begin{figure}[htbp]
		\centering
		\includegraphics[width=0.9\linewidth]{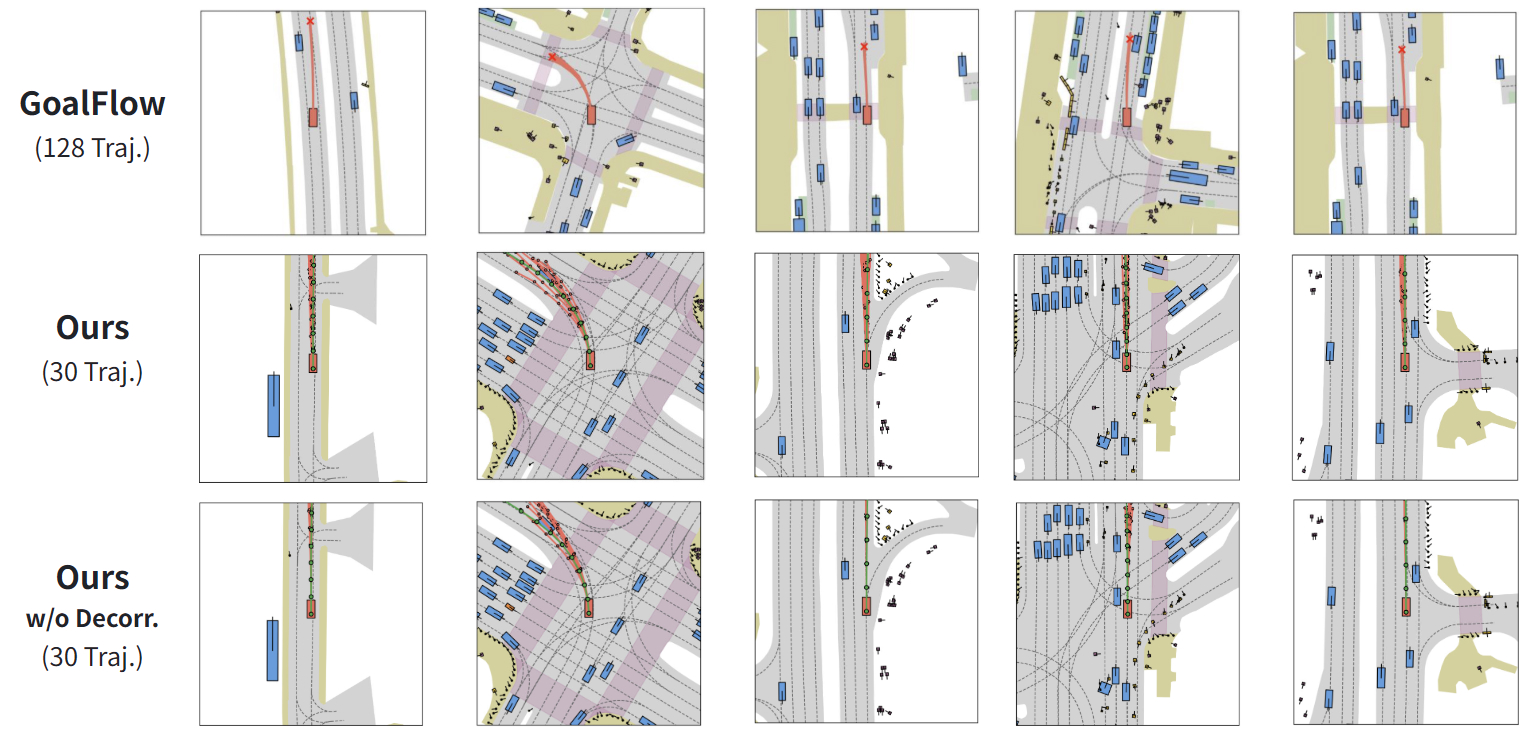}
		\caption{Visualization on multi-mode diverse trajectories.}
		\label{fig:vis_multi}
\end{figure}




\section{Conclusion}
\label{sec:conclusion}
In this work, we propose TransDiffuser, an encoder-decoder based generative trajectory model for end-to-end autonomous driving.
The encoded scene features and motion features serves as conditional input of the Diffusion-based denoising decoder.
We introduce a simple yet effective multi-modal representation decorrelation optimization mechanism to encourage more diverse trajectories from the continuous action space. 
Experiments on the planning-oriented NAVSIM benchmark demonstrate its superiority in generating high-quality diverse trajectories, without any trajectory anchors or scene priors.
For future works, considering more real-world scenarios, the vehicle should generate multiple planning trajectories better aligned with human driver commands or styles, these attempts require reinforcement learning based preference optimization techniques \cite{trajhf,breaking} and vision-language-action model architectures \cite{openvla,emma} to achieve the better balance between diversity and safety, which is a crucial concern for deployment. We also plan to explore vision-only generative approaches on NAVSIMv2 \cite{navsimv2} for open-looped evaluation.

\section*{ACKNOWLEDGMENT}
This work is a follow-up work of our team's previous proposed TrajHF \cite{trajhf}. 
We appreciate the valuable feedback and advice from Tianyu Li from OpenDrive Lab and Hangjie Mo from Hefei University of Technology.
We also appreciate the assistance of Jianwei Ren from Shanghai Qi Zhi Institute, along with Yue Wang, Chuan Tang and other researchers and engineers from LiAuto for TrajHF re-implementation and other technical supports.










\bibliographystyle{IEEEtran}
\bibliography{IEEEabrv,example}

\end{document}